%% file: Feldman_socialsense_2021_static_bib.tex
\def\year{2021}\relax
\documentclass[letterpaper]{article} % DO NOT CHANGE THIS
\usepackage{aaai21}  % DO NOT CHANGE THIS
\usepackage{times}  % DO NOT CHANGE THIS
\usepackage{helvet} % DO NOT CHANGE THIS
\usepackage{courier}  % DO NOT CHANGE THIS
\usepackage[hyphens]{url}  % DO NOT CHANGE THIS
\usepackage{graphicx} % DO NOT CHANGE THIS
\usepackage{natbib}  % DO NOT CHANGE THIS AND DO NOT ADD ANY OPTIONS TO IT
\usepackage{caption} % DO NOT CHANGE THIS AND DO NOT ADD ANY OPTIONS TO IT
\usepackage{csquotes}
\usepackage{booktabs}
\usepackage{comment}
\usepackage[hang,flushmargin]{footmisc}
\usepackage{paralist}
\usepackage[dvipsnames]{xcolor}
\definecolor{cyan}{RGB}{0, 128, 128}
\definecolor{green}{RGB}{0, 128, 0}
\definecolor{blue}{RGB}{0, 0, 128}

\title{Analyzing COVID-19 Tweets with Transformer-based Language Models} %JF: I added a line break here to keep each idea on the title in its own line
\author {
	Philip Feldman\textsuperscript{\rm 1, 2}
	Sim Tiwari \textsuperscript{\rm 2}
	Charissa S. L. Cheah \textsuperscript{\rm 2}
	James R. Foulds \textsuperscript{\rm 2}
	Shimei Pan \textsuperscript{\rm 2} \\
}
\affiliations {
	\textsuperscript{\rm 1} ASRC Federal, Beltsville, Maryland, USA \\
	\textsuperscript{\rm 2} University of Maryland Baltimore County, Baltimore, Maryland USA \\
	\textsuperscript{\rm 1}philip.feldman@asrcfederal.com, \textsuperscript{\rm 2}\{simt1, ccheah, jfoulds, shimei\}@umbc.edu  
}

\begin{document}
    % Have final version by COB Wednesday for review
    % Reviews Thursday to Friday morning
    % Roll in changes by the 19th
    % DON"T FORGET TO MAKE IT A SINGLE FILE!!!
	
	\maketitle
	
	\begin{abstract}
		This paper describes a method for using Transformer-based Language Models (TLMs) to understand public opinion from social media posts. In this approach, we train a set of GPT models on several COVID-19 tweet corpora that reflect populations of users with distinctive views. We then use prompt-based queries to probe these models  to reveal insights into the biases and opinions of the users. We demonstrate how this approach can be used to produce  results which resemble polling the public on diverse social, political and public health issues. The results on the COVID-19 tweet data show that transformer language models are promising tools that can help us understand public opinions on social media at scale.
	\end{abstract}
	
	\noindent 
	\input{text/introduction}

	\input{text/literature}

	\input{text/dataset}
	\input{text/methods}

	\input{text/results}
	\input{text/discussion}

	\vspace{.1cm}
	\hrule
	\vspace{.1cm}
	\noindent
	\textit{This paper is based upon work supported by the National Science Foundation under grant no. 2024124.}
\end{document}

%% file: text/introduction.tex
\section{Introduction}
\label{subsec:big_data_agents}

Large-scale research based on feedback from humans is difficult, and often relies on labor-intensive mechanisms such as polling, where statistically representative populations will be surveyed using phone interviews, web surveys, and mixed-mode techniques. Often, for longitudinal studies, participants in a survey may need to be recontacted to update responses as a result of changing environments and events~\cite{fowler2013survey}. 

As social media has become ubiquitous, many attempts have been made to determine public opinion by mining large amount of social media data, often spanning years, which is available from online providers such as Twitter, Facebook and Reddit, e.g.~\cite{colleoni2014echo, sloan2015tweets}.  Though social data can be mined in a variety of ways, answers to specific questions frequently can not be obtained without expensive manual coding.

This may be ready to change with the emergence of large transformer-based language models (TLMs) like the  GPT series~\cite{radford2018improving} and BERT~\cite{devlin2018bert}. These models are trained on massive text datasets such as BookCorpus, WebText and Wikipedia. They implement a transformer-based deep neural network architecture which uses attention to allow the model to selectively focus on the segments of the input text that are most useful in predicting target word tokens. A pre-trained GPT model can be used for generating texts as a function of the model and a sequence of word tokens provided by users.  We call the sequence of words provided by users a \enquote{prompt} or a \enquote{probe}, which is specifically designed to set up the theme/context for GPT to generate sentences. Since the model is not trained using any hand-crafted language rules, it effectively learns to generate natural language by observing a large amount of text data. In doing so, it captures semantic, syntactic, discourse and even pragmatic regularities in language. GPT models were shown to generate text outputs often indistinguishable from that of humans~\cite{floridi2020gpt}. 

As such, these models contain tremendous amounts of information that can be used to answer questions about the content and knowledge encoded in the training text. Unfortunately, the knowledge captured in these TLMs is latent (e.g., millions of neural network model parameters),  can be  difficult to interpret. In this study, by using carefully constructed prompts/probes, we \enquote{poll} the model to gain access to the latent knowledge contained in the model in ways that are analogous to accessing the knowledge in a population by surveying a random sample  of individuals.  

\begin{figure}[h]
	\centering
	\fbox{\includegraphics[width=12em]{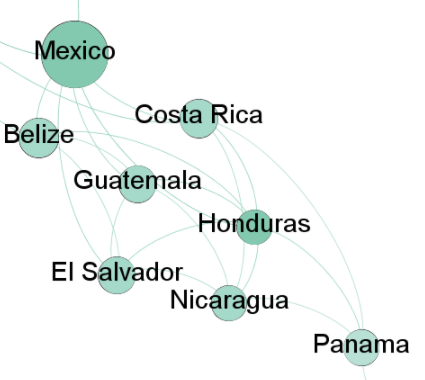}}
	\caption{\label{fig:gpt3_central_america} Central America reconstruction.}
\end{figure}

We can illustrate this process against a known ground truth (the spatial relationships between countries) by polling the GPT-3\footnote{GPT-3 accessed using the OpenAI beta API} repeatedly with the prompt \enquote{A short list of countries that are nearest to \rule{5mm}{0.15mm}, separated by commas:}. The prompt is initially seeded with a value such as \enquote{United States}. Recursive responses create a graph, and a force-directed layout approximately reconstructs the original country relationships\footnote{Full map at tinyurl.com/gptworldmap} (Figure~\ref{fig:gpt3_central_america}).  

Using this approach, we can create \enquote{maps} that approximate the local spatial relationships of $\approx$ 75\% of current countries. The presence or absence of a country is correlated with population. This appears to be a bias in the model.

In this paper, we employ a similar approach that combines a repeated-prompt based query method with data analysis and visualization to analyze the beliefs, biases and opinions of Twitter users in the context of the COVID-19 pandemic. We describe our method for GPT model training and fine tuning, determining probes, and analyzing and visualizing the output from the model. Lastly, we discuss its limitations and implications for this approach to computational sociology.

%% file: text/literature.tex
\section{Related Work}
%In this section, we review our related work in three categories.
Since the introduction of the transformer model in 2017, TLMs have become a field of study in themselves. The transformer 
%is unlike perceptrons and convolutional neural networks in that it  uses no pre-defined objective functions~\cite{rogers2020primer}. Rather it
uses self attention, where the model computes its own representation of its input and output~\cite{vaswani2017attention}. So far, significant research has been in increasing the performance of these models, particularly as these systems scale into the billions of parameters, e.g. ~\cite{radford2019language}. Among them, BERT~\cite{devlin2018bert} and GPT~\cite{radford2018improving} are two of the most well known TLMs used widely in boosting the performance of diverse NLP applications.

Understanding how and what kind of knowledge is stored in all those parameters is becoming a sub-field in the study of TLMs. %Since language models require no human supervision to train, do not have schemas like traditional databases, and can be queried using natural language, they make an attractive potential mechanism for storing and retrieving information. 
Among them, ~\cite{petroni2019language} used probes that present a query to the mode as a cloze statement, where the model fills in a blank (e.g. \enquote{Twinkle twinkle \rule{.9cm}{0.15mm} star}).
%They found that BERT-Large was \enquote{competitive with non-neural and supervised alternatives.}
Research is also being done on the creation of effective prompts. Published results show that mining-based and paraphrasing approaches can increase effectiveness in masked BERT prompts over manually created prompts~\cite{jiang2020can}. 
%For example, mined prompts can be produced by mining phrases in the Wikipedia corpus that can be generalized as template questions such as \textit{x was born in y} and \textit{capital of x is y}. These can then be filled in using sets of subject-object pairs. Improvements using this technique can be substantial, with improvements of 60\% over manual prompts. Paraphrasing, or the simplification of a prompt using techniques such as back-translation can enhance these results further~\cite{jiang2020can}. 
In another study using GPT models fine-tuned on descriptions of chess games, it was shown that models trained on a corpora of approximately 23,000 chess games accurately replicated human gameplay patterns~\cite{feldman2020navigating}. 
%Statistical analysis comparing the spectral characteristics of human (ground truth) and synthesized games were found to be statistically similar with a $> 97\%$ probability. This indicates that TLMs accurately reflect the biases and beliefs in the corpora they have been trained on. 

Using TLMs to evaluate social data is still nascent. A study by \cite{palakodety122020mining} used BERT fine tuned on YouTube comments to gain insight into community perception of the 2019 Indian election. 
%They created weekly corpora of comments and constructed a tracking poll based on the prompts \enquote{Vote for MASK} and \enquote{MASK will win} and then compared the probabilities for the tokens for the parties BJP/CONGRESS and candidates MODI/RAHUL. The results substantially tracked traditional polling.

Lastly, we cannot ignore the potential weaponization of TLMs.   OpenAI has shown that the GPT-3 can be \enquote{primed} using \enquote{few-shot learning}~\cite{brown2020language}.~\cite{mcguffie2020radicalization} primed the GPT-3 using mass-shooter manifestos with chilling results.

%% file: text/dataset.tex
\section{Dataset}
\label{subsec:data_collection}
In this study, we investigated the feasibility of employing GPT to analyze COVID-19 related tweets. Using the Twitter Search API, we collected tweets from the USA if they included at least one of the identified keywords/hashtags. As of this writing, the list of case-insensitive keywords/hashtags include \enquote{coronavirus}, \enquote{covid19}, \enquote{sars-cov-2}, \enquote{pandemic}, \enquote{chinavirus}, \enquote{social distancing}, \enquote{mask}, \enquote{ventilator}, \enquote{shelter in place} etc. Terms are reevaluated and updated monthly.
%\jimmy{Were the tweets lowercased when matching the keywords?}

%The keywords/hashtags were first identified by observing Twitter’s \enquote{trending} information as well as Google Trends’ keyword suggestions. To ensure that the identified keywords/hashtags would result in tweets that are highly relevant to the COVID pandemic, a manual precision test was performed based on 50 sample tweets retrieved using each keyword/hashtag. Those with precision $\ge 80\%$ were selected to be used in the final API queries. 

%To maximize the tweets we could retrieve based on our subscription, we used the Premium 30-day search API to retrieve tweets month by month starting from March 2020. We also used the Premium Archival search API to retrieve Tweets that are more than 30 days old. Due to the volume limit posted by the Twitter search APIs, we were only able to retrieve 40\% of the tweets satisfying the query criteria.  
So far, we have retrieved a total of 18,703,707 tweets from Nov. 2019 to the time of this writing. %All retrieved tweets are in JSON format and have been parsed and stored in a MySQL database.
For this research, we constructed three datasets to train three separate GPT models based on three COVID-19 keywords: \textit{covid, sars-cov-2}, and \textit{chinavirus}.
The most common term by Twitter users to refer to the disease, was  \textit{covid}, and was thought to represent the perspectives of the general public;  \textit{sars-cov-2} was chosen for its greater use in science-related contexts; while \textit{chinavirus} was chosen for its arguably racist connotations.
%we attempted to find terms that were used regularly throughout the pandemic and reflected usage by different groups. We chose three case-insensitive tags: \textit{chinavirus, covid}, and \textit{sars-cov-2}. Note that these tags could be in isolation, present within hashtags, or part of larger terms such as COVID-19. The tag \textit{chinavirus} was chosen for its association with President Trump and possible racist connotations. \textit{Covid} was chosen for its popularity, and was thought to represent the perspectives of the broader public. Lastly, the tag \textit{sars-cov-2} was chosen based of its greater use in science-based online documents and social media.

Table~\ref{tab:tweet_counts} shows the size of each dataset.  
%\jimmy{I guess the sentences on the very different sizes of the dataset were removed for space, but I think it's important to say something about that here. }
As one would expect from its more common usage, we collected approximately 350 times more tweets associated with the \textit{covid} than the other two tags. This would have some implications in the behavior of the trained models, which we will describe in the fine-tuning section. Please note the three datasets may be overlapping (e.g.,  a tweet tagged with both \textit{chinavirus} and \textit{covid}). There are also many pandemic tweets not in any of the datasets (e.g., those tagged with \textit{social distance}.) 

\begin{table}[h!]
	\begin{center}
		\caption{Tweet counts in each dataset}
		\label{tab:tweet_counts}
		\begin{tabular}{ccc}
			\toprule
			chinavirus&  covid& sars-cov-2 \\
			\midrule
			14,950&  7,015,582&  25,768\\
			\bottomrule
		\end{tabular}
	\end{center}
\end{table}

We show three tweets from each dataset. These tweets were posted at the beginning of April 2020, when the pandemic's first wave was starting to happen in the USA. 

\begin{displayquote}
\textbf{chinavirus dataset}: \textit{Just to remind everyone why social healthcare is terrible...the UK has a 7\% mortality rate of \#ChinaVirus \#WuhanVirus \#coronavirus while America is currently hovering around 1.7\% of reported cases}
\end{displayquote}

\begin{displayquote}
\textbf{covid dataset}: \textit{This can't be happening. These are our first responders! If they aren't protected our entire country is at risk.  These doctors and nurses are on the front lines risking their life's.  With a RAIN PANCHO?! \#coronavirus \#COVID \#COVID2019 https://t.co/KWPVGuIgPW}
\end{displayquote}

\begin{displayquote}
\textbf{sars-cov-2 dataset}: \textit{Currently, incubation period of coronavirus, SARS-CoV-2, is considered to be 14 days. Health authorities across countries are adopting a 14-day quarantine period, based @WHO guidelines.}
\end{displayquote}

These samples suggest  qualitatively differing perspectives: the \textit{chinavirus} appears reactionary and nationalistic; the \textit{covid} tweet is more emotional; while the \textit{sars-cov-2} tweet is detailed and explicit.  

%% file: text/methods.tex
\section{Methods}

%\shimei{Since the performance of the chess model may not be directly linked to the performance of these three twitter models, a better place for the above paragraph may be in the related work section. }

In this section, we describe how we customize a pre-trained GPT model with COVID-19 tweets and how we design prompts to setup the context and probe the models to reveal the answers to our questions. 

\subsection{Model Training and Fine Tuning}
For this research, we employ the pre-trained GPT-2 model with 12-layers, 768-hidden, 12-heads, 117M parameters hosted by Huggingface.com\footnote{huggingface.co/gpt2}. 
%Three corpora were created by selecting all tweets in the database that had text which matched our \textit{chinavirus}, \textit{covid}, and \textit{sars-cov-2} tags. The text files created for the training and evaluation datasets, had 25\% of the tweets were randomly allocated to the evaluation text file. Using the Huggingface API\footnote{github.com/huggingface/transformers/tree/master/examples/language-modeling}, we then retrained the original model on these texts. 
We then fine tune the pre-trained GPT-2 model using the three  COVID-19 tweet datasets described above to produce three GPT models, one for each dataset. We did not use the latest GPT-3 model as it does not allow us to retrain it with new data. 

To demonstrate the behavior of these models, we use the beginning of the same tweets that are shown in the previous section as the prompts to generated three synthetic tweets. In each generated tweet, the prompt is shown in the brackets, and the generated content is shown in \textit{italics}:

%\shimei{I didn't see any prompts used in generating the following tweets. Please check if my interpretation is correct}

%\phil{The prompts are the first parts of the tweets from the previous section:}
%\begin{itemize}
 %   \item \phil{Just to remind everyone why social healthcare is}
%    \item \phil{This can't be happening, These are our first responders!}
%    \item \phil{Hmmmm... I thought the virus was SARS-CoV-2.}
%\end{itemize}
%\phil{I guess I need to say this more clearly}
%\shimei{If that's the case, then these are not very good examples.  For example, in tweet 1 and 2 above, the system mostly generated a few hashtags. Is there any better tweets we can used to replace tweet 1 and 2. If not, since it is unlikely the system can reproduce the same tweet given the same prompt (depending on how representative the original tweet is in the training data), why not we just choose three Representative generated tweets without prompts?}

%Here are three tweets randomly generated by the three GPT models with only a start symbol as the prompt:

\begin{displayquote}
	\textbf{Chinavirus model:} [Just to remind everyone why social healthcare] \textit{doesn't work. Stop being afraid. @realDonaldTrump @FLOTUS https://t.co/lFh9R1KdU}
\end{displayquote}

\begin{displayquote}
	\textbf{covid model:} [This can't be happening. These are our first responders!] \textit{Please keep them in your thoughts and prayers! \#NursesCOVID19 https://t.co/K8xnCfGfjC}
\end{displayquote}
 
\begin{displayquote}
	\textbf{Sars-cov-2 model:} [Currently, incubation period of coronavirus, SARS-CoV-2, is considered to be 14 days.] \textit{The duration of incubation is 12-14 days and the virus can persist for up to 18 hours (depending on temperature and humidity).}
\end{displayquote}

We can see that each model has captured not only the format of tweets (e.g., using mentions such as @realDonaldTrump, @FLOTUS or short URLs such as https://t.co/lFh9R1KdU) but also subjective and semantic aspects of the original training data. As we can see, the models are not \textit{reconstructing} the tweets. Rather, they are stochastically generating new text based on the prompts and latent knowledge in the model. This property allows us to obtain responses to our queries which may reflect the beliefs  typical of the set of twitter users in the training data.

\subsection{Prompt design}

%\shimei{The above paragraph is a topic better used in the discussion section. In general, we want to present our method without raising all kinds of questions. Then in the discussion section, we discuss the limitations/weakness of the proposed method. }
Prompts can be regarded as a form of conditioning or training, as discussed by \cite{brown2020language}. The length and content of the probe can be structured so that a desired response is more likely. 

Finding probes that resulted in useful responses was an interactive process, where we would develop ideas for probes and try them out on the models in small batches.
%using a small program that would load a model and generate a small number of responses, typically 10 - 20 in response to the probe. 
For example, the probe \enquote{Dr. Fauci is}  allows the model to produce adjectives (\enquote{likable}) adverbs (\enquote{very}), determiners (\enquote{a}), and verbs (\enquote{running}) as the next token. Changing the probe to \enquote{Dr. Fauci is a} constrains the next word to more likely to be an adjective or a noun. If we use the next nouns or adjectives after the prompts as the responses to our inquiries, the probe \enquote{Dr. Fauci is a} may produce more direct answers.
%\shimei{Are we talking about the next token here ? If we are talking about the next tokens as the responses to our inquiry, in the following samples, we should just focus on the next tokens and compare their differences. The current examples do not seem to be connect with the above paragraph as I do not see a contrast between these two models. Both generate reasonable and plausible responses wrt Dr. Fauci  }

Example output from the \textit{covid} model is shown in Table~\ref{tab:probe_eval}. %Figure~\ref{fig:fauci}.

\begin{table}[h!]
	\begin{center}
		\caption{Similar probes and different GPT outputs. We bold face the words if the first nouns are extracted as the answers }
		\label{tab:probe_eval}
		\begin{tabular}{ll}
			\toprule
			Dr. Fauci is & Dr. Fauci is a  \\
			\midrule
			out of the \textbf{spotlight} &  \textbf{hero} in the war against COVID\\
			at it again & dangerous \textbf{man}.\\
			100\% correct & medical \textbf{genius}\\
			on top of \textbf{everything} & \textbf{liar}. It was never about COVID19\\
			\bottomrule
		\end{tabular}
	\end{center}
\end{table}
%\shimei{suggest to replace "Trump supporter" with another example as I mentioned earlier, we will discuss all the weaknesses/limitations in the discussion section. You do not want to make people question the validity of the method throughout the entire paper}
\begin{comment}
\begin{figure}[h]
	\centering
	\fbox{\includegraphics[width=22em]{figures/Fauci}}
	\caption{\label{fig:fauci} Probe evaluation example}
\end{figure}
\end{comment}

For these relatively small models, we found that shorter probes, typically 3 - 7 words would produce useful results. By varying the prompts, we could tune the results to explore   the latent knowledge captured by GPT.

Finally, we designed a set of prompts to probe the GPT model to reveal the Twitter public's opinion/attitude toward various COVID-19 pandemic-related topics such as: whom to blame for the pandemic? How do people feel during the pandemic? Is there any systematic bias towards certain demographic groups?  For example, the prompt \enquote{During the pandemic, [xxx] have been feeling} would be filled in with the terms [Americans, Asians, Black folks] to compare the feelings of different people during the pandemic.

\subsection{Response/Answer Extraction}
There are multiple ways to extract answers to different queries from the GPT model. First, directly from the model. Transformer language models such as the GPT generate the next token in a sequence based on a probabilistic distribution computed by the model. We can directly use the output probability of a word given a prompt as the probability related to an answer. For example, given the prompt  \enquote{Dr. Fauci is a}, the model can directly output the probability of the word \enquote{scientist} or \enquote{liar} appearing as the next word.
%\jimmy{expert would require "an" instead of "a," so maybe pick a different word which doesn't start with a vowel}
Second, extracting answers based on output sample analysis. For each prompt, we can generate a large number of representative tweets using the model. We can then compute statistics based on the generated samples. In this preliminary study, we adopted the second approach.  We will explore the first approach in future work.

%This is because the tokenizer used by GPT frequently breaks down words into  subword units (e.g, instead of keeping the word \enquote{Coronavirus} as one unit, the GPT breaks it down to multiple tokens such as \enquote{Cor}, \enquote{on}, \enquote{av} and \enquote{irus}. Since the GPT only directly outputs the probability associated with each token, frequently we cannot directly output the probability of a word or phrase  without additional statistical analysis.\jimmy{I don't know if you want to belabor this point, since we could actually have solved this with a bit more work, by considering multiple tokens up to a space or punctuation.} 

Specifically, for each prompt, each model generated 1,000 sample responses. Each response was stored in a database, along with parts-of-speech tags and a sentiment label (either positive or negative) automatically assigned by a sentiment analyzer~\cite{akbik2019flair}. 
%The data for each run is stored in a MySql relational database along with meta information about the experiment for later analysis.
\begin{comment}
%For the second dataset, we used the top terms from each probe from the first dataset to build a list of tokens that we could examine the rank likelihood for each token generated by the model.

%Autoencoding Transformer models such as the GPT generate the next token in a sequence based on a probabilistic function. With temperature set to zero, the token with the highest probability will be selected. Higher temperatures allow for other, lower ranked tokens to be selected, altering how the text sequence develops. This means that the probability of all the tokens in the model is calculated at each step in the process. To extract this rank information, we used the The Ecco library\footnote{github.com/jalammar/ecco}. Token-by-token interrogation of the model is computationally expensive, so for this approach, only 10 responses were generated per model per probe. 
\end{comment}
Statistics were computed from the samples, focusing on the relationships between probes and models. Initial statistics were gathered on the first words directly after each prompt in each response as they were generated directly from the given probe, and are most likely to vary with and impacted by the probe. Since next word analysis is likely to produce high probable functional words that do not carry specific meaning such as \enquote{the} and \enquote{has}, parts-of-speech tags were used to extract the first noun (\enquote{virus}) or noun-noun combination (\enquote{chinese virus}), or adjective (\enquote{anxious}). Lastly, we computed the percentage of positive tweets in the 1000 samples generated per prompt per model.

One might ask why should we analyze synthetic tweets generated by GPT  instead of the real tweets directly? For some insight, we can look at Table~\ref{tab:probe_freq} which shows the number of times each probe appears exactly in the 18 million real tweets we retrieved:

\begin{table}[h!]
	\begin{center}
		\caption{Probe frequency in 18 million COVID  tweets}
		\label{tab:probe_freq}
		\begin{tabular}{lr}
			\toprule
			Probe & Count  \\
			\midrule
			Donald Trump is a & 1,423\\
			Dr. Fauci is a &  427\\
			The pandemic was caused by & 6\\
			For the pandemic, we blame & 0\\
			\bottomrule
		\end{tabular}
	\end{center}
\end{table}

Though there are enough results  to perform statistical analysis related to Donald Trump and Dr. Fauci, the data is not sufficient to support an analysis with statistical significance for the COVID-19 causes and blames. Because transformer language models such as the GPT create tweets based on statistically valid patterns~\cite{feldman2020navigating}, each synthetic tweet may encode information derived from a large number of real tweets that are syntactically and semantically related. Further, in principle, the models can generate unlimited  datasets to facilitate robust data analysis.

%% file: text/results.tex
\section{Preliminary Results}
In this section, we present results that explore biases in the tweet corpora that each model was trained on. %Because these models have been trained from data that contains biases in tweets that are associated with the tags \textit{chinavirus}, \textit{covid}, and \textit{sars-cov-2}

\subsection{Polling the General Public on Twitter}
%\jimmy{Mismatched capitalization in heading}

The \textit{covid} model was trained on the most data and more representative of mainstream Twitter users' opinion/attitude. As such, it was more able to provide more granular responses to our probes. Our first set of results will focus on the behavior of the \textit{covid} model.

\subsubsection{PROBE: \enquote{The pandemic was caused by}}
Based on the normalized count (i.e. percentage) of the first nouns to appear after this probe in the 1000 generated samples, we list the top ranked nouns in Table~\ref{tab:pandemic_caused_by}. 
\begin{comment}
\begin{figure}[h]
	\centering
	\fbox{\includegraphics[width=22em]{figures/pandemic_caused_by}}
	\caption{\label{fig:pandemic_caused_by} The pandemic was caused by: top responses}
\end{figure}
\end{comment}
\begin{table}[h!]
	\begin{center}
		\caption{\enquote{The pandemic was caused by} top responses}
		\label{tab:pandemic_caused_by}
		\begin{tabular}{lr}
			\toprule
			Cause & Percentage  \\
			\midrule
			Coronavirus &  31\% \\
			Virus & 11\% \\
			China & 6\% \\
			Lab & 2\% \\
			Accident & 2\% \\
			\bottomrule
		\end{tabular}
	\end{center}
\end{table}

The vast majority of the responses (31\%) attribute the coronavirus explicitly, while an additional 11\% referred to viruses in general. Typical tweets generated by the model after the prompt \enquote{The pandemic was caused by} include  \textit{a novel \textbf{coronavirus} known as SARS-CoV-2} and \textit{a \textbf{virus} known as SARS-CoV-2}. Variations of these statements make up over half of the generated results ranging from fatalistic, \textit{a \textbf{virus} that could not have been contained} to conspiratorial, \textit{ a novel \textbf{coronavirus} that originated in a lab in Wuhan}.

The next values appear to be more focused on human causes. For example,  \enquote{caused by} \textit{\textbf{China}, which had unleashed COVID19 on the world and is responsible}. Further down this list align with conspiracy theories, where \enquote{the pandemic was caused by} \textit{a \textbf{lab} accident. You can bet there were dozens of other deaths}, and \enquote{caused by} \textit{a 'genetic \textbf{accident}' of the genes of aborted babies}.

\subsubsection{PROBES: Blame for the pandemic}
To see if the models would distinguish between \textit{cause} and \textit{blame}, we tried the probe \enquote{For the pandemic, we blame}. The most common response was to blame  President Trump (Table~\ref{tab:pandemic_blame}). 
\begin{comment}
\begin{figure}[h]
	\centering
	\fbox{\includegraphics[width=22em]{figures/pandemic_blame}}
	\caption{\label{fig:pandemic_blame} covid model: \enquote{For the pandemic, we blame}}
\end{figure}
\end{comment}
\begin{table}[h!]
	\begin{center}
		\caption{\enquote{For the pandemic, we blame} top responses}
		\label{tab:pandemic_blame}
		\begin{tabular}{lr}
			\toprule
			Cause & Percentage  \\
			\midrule
			Trump &  35\% \\
			China & 13.5\% \\
			COVID-19 & 5.7\% \\
			Media & 5.3\% \\
			Government & 2.6\% \\
			\bottomrule
		\end{tabular}
	\end{center}
\end{table}
%\shimei{can we have top 5 instead of 4 in the above example, just to be consistent?}
Responses such as this one are common: \enquote{For the pandemic, we blame} \textit{\textbf{Trump} for the catastrophic response}. Tweets that blamed the media often blamed the government as well: \enquote{we blame} \textit{the \textbf{Media} and \textbf{Government} for not telling the truth}.

\subsubsection{PROBES: How distinct groups are feeling}
To extract the public opinions about ethnic groups, we ran three probes, each of which began with \enquote{During the pandemic,}, and finished with: 
\begin{inparaenum}[1)]
	\item \enquote{Asians have been feeling}
	\item \enquote{Black folk have been feeling}
	\item \enquote{Americans have been feeling.}
\end{inparaenum}

\begin{figure}[h]
	\centering
	\fbox{\includegraphics[width=22em]{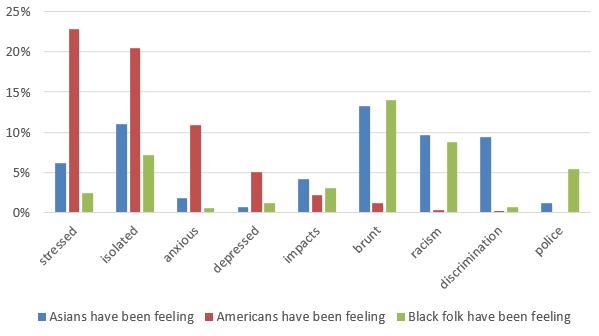}}
	\caption{\label{fig:asian_black_americans_feeling} Asian/Black/American feeling, top responses. X-axis is sorted by the normalized frequency (percentage) of each word appearing in the tweets generated by the \textit{covid} model with \textit{Americans have been feeling} as the prompt.}
\end{figure}

The top results are shown in Figures~\ref{fig:asian_black_americans_feeling}. Common among the Asian and Black groups is the term \enquote{brunt}. In the Asian results, tweets like \enquote{During the pandemic, Asian Americans have been feeling \textit{the \textbf{brunt} of discrimination and harassment.}} are common. Alternatively, Black results emphasize \enquote{feeling \textit{the \textbf{brunt} of racism, discrimination, and police brutality}}. All groups have substantial numbers of responses that refer to isolation, with output like \enquote{have been feeling \textit{\textbf{isolated}, lonely \& disconnected.}} However, the dominant term in the American responses is \enquote{stressed}\footnote{for this analysis the counts for \enquote{stressed}, \enquote{stress}, and \enquote{strain} were combined}. For example, \enquote{Americans have been feeling \textit{a lot more \textbf{stressed} and anxious about a new normal}}, and \enquote{\textit{\textbf{anxious}, stressed, hopeless, and depressed}}. These generated texts indicate that the model is presenting a more subjective feeling set of responses for Americans in general, while ethnic sub-groups are feeling the brunt of external forces.

Please note that the above results are about \enquote{how groups feel} based on the Twitter general public. To poll the feeling of each ethnic group directly, we would need to use a prompt like \enquote{I am an Asian. I have been feeling}. 
%\jimmy{Use `` and '' to get quotation marks the right way around in LaTeX}

%\shimei{"stressed" is not a noun. How did you extract stressed? So, you also extract adjectives in addition to nouns?}
%\phil{each sentence is stored with its parts of speech component by word. I'm currently extracting Nouns (NN), and Adjectives(JJ) because it works better with the 'feeling' probe}

%\input{text/how_we_feel}

\subsection{Polling Different Populations}
In addition to the \enquote{covid} model, we created models trained on tweets containing the \enquote{chinavirus} and \enquote{sars-cov-2} tags. In this section, we compare the outputs across the three models, similar to polling to different sub-populations on Twitter.  Each model generated 1,000 synthetic tweets for each probe, allowing for direct comparison between the generated data. 

\subsubsection{Donald Trump and Dr. Fauci}
\begin{figure}[h]
	\centering
	\fbox{\includegraphics[width=22em]{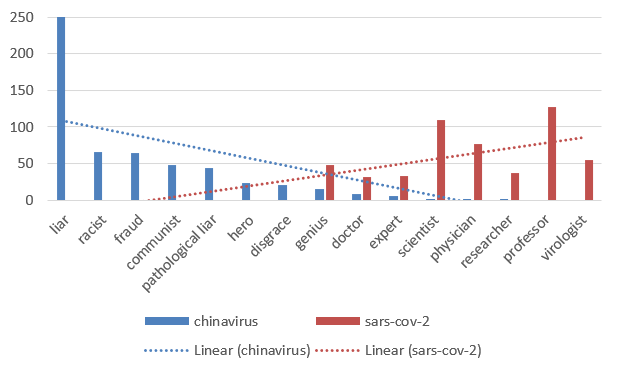}}
	\caption{\label{fig:nouns_fauci_trump} Fauci nouns extracted from different models, where the x-axis is sorted by their frequency computed from the Twitter samples synthesized by the \enquote{chinavirus} model.}
\end{figure}

One of the most polarizing prompts that we found was \enquote{Dr. Fauci is a}\footnote{Thanks to Dr. Peterson of Montgomery College for the suggestion}. This created distinct sets of responses for each of the models, as seen in Figure~\ref{fig:nouns_fauci_trump}. The chinavirus model produced tweets such as \enquote{Dr. Fauci is a \textit{\textbf{liar} and a demagogue \#ChinaVirus}.} Sorting by term frequency based on the outputs of this model produces an opposed trend in the sars-cov-2 model. Linear regression on each model's term frequency clearly shows this interaction. The dominant terms produced by the sars-cov-2 model for this prompt are \textit{professor, scientist}, and \textit{physician}. The generated content uses a more informational style, such as \enquote{Dr. Fauci is a \textit{\textbf{professor} and \textbf{physician}. He authored and co-authored several papers published on SARS-CoV-2}}

To examine the differences in emotional tone that these models produced with the \enquote{Dr. Fauci is a} probe, We ran an additional probe, \enquote{Donald Trump is a}, and compared the sentiment of the tweets synthesized from each probe across all three models, using an existing sentiment analyzer  \cite{akbik2019flair}. This is shown in Table~\ref{tab:sentiment_fauci_trump}:

\begin{table}[h!]
	\begin{center}
		\caption{Trump / Fauci Positive Sentiment}
		\label{tab:sentiment_fauci_trump}
		\begin{tabular}{lccc}
			\toprule
			Probe &chinavirus&  covid& sars-cov-2 \\
			\midrule
			Dr. Fauci is a& 13.3\%&  33.1\%&  53.6\%\\
			Donald Trump is a& 44.4\%&  28.4\%&  27.1\%\\
			\bottomrule
		\end{tabular}
	\end{center}
\end{table}

We see a similar pattern to that seen in Figure~\ref{fig:nouns_fauci_trump}. In the response to the \enquote{Dr. Fauci is a} probe, the chinavirus model generates only 13\% positive responses, while it generates approximately 45\% positive text in response to \enquote{Donald Trump is a} (e.g. here is one such tweet produced by the model:  \enquote{\textbf{Donald Trump is a} \textit{great politician and a man of integrity}.}). The covid model falls between the other two models, particularly with respect to the Dr. Fauci probe. It is not significantly different from the behavior of the sars-cov-2 model in response to the Donald Trump probe.

\subsection{Polling Over Time}
The proposed method can also be used to poll opinions at different times. We can have two different ways to poll the model over time. First, fine tune each model with time sensitive data (e.g., fine tune the model with pre-pandemic versus during and post pandemic data); Second, we may use time-sensitive prompts such as \enquote{in March, 2020, Americans have been feeling}. In this study, we perform a coarse pre- and during-pandemic analysis. We therefore used the first method.

\begin{figure}[h]
	\centering
	\fbox{\includegraphics[width=23em]{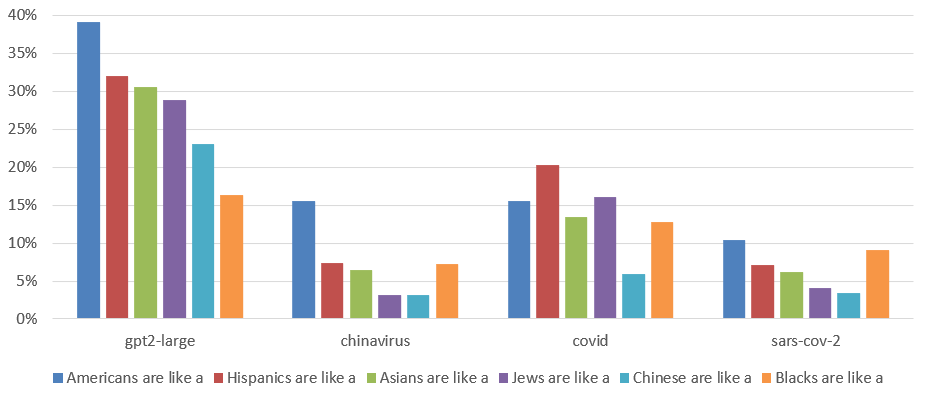}}
	\caption{\label{fig:racemetaphor} Public sentiment towards different demographic groups before and during the pandemic. The y-axis is the percentage of positive sentiment associated with the samples generated with each prompt by each model.}
\end{figure}

As shown in Figure~\ref{fig:racemetaphor}, we summarize the public sentiment towards different demographic groups before and after the pandemic. The GPT2-large model on the left was trained on general web data before the pandemic while the other three models were fine-tuned with the twitter covid data we collected during the entire course of the pandemic to date. The probes used in the analysis include \enquote{[xxx] are like a} where [xxx] can be Americans, Hispanics, Asians, Blacks, Jews and Chinese. We use the GPT's responses to these metaphors to assess the public sentiment towards different  demographic groups.  

As shown in the chart, the sentiment in general is much more positive before (outputs from 2019 GPT2-large) than during the pandemic (outputs from all three pandemic-related models). This is true across all demographic groups we considered in the experiment.  Moreover, before the pandemic, the sentiment towards \enquote{Americans} is the most positive  while that towards "Blacks" is the most negative. During the pandemic, the sentiment towards Chinese has turned decisively negative. This is true across all three pandemic models. Inspecting tweets generated by different models towards the Chinese, the GPT2-large model generates tweets like \enquote{\textbf{We  think  Chinese  are  like}  \textit{a  lot  of  other groups - very loyal to their own, have great energy}}. However, during the pandemic,  the chinavirus model generates tweets like \enquote{\textbf{We think Chinese are like} \textit{a snake eating bats in a cauldron. \#ChinaVirus}}. 
   
\begin{comment}
Blacks by model
\begin{itemize}
    \item \textbf{GPT-Large}: \textit{We think blacks are like a species, and black people have different needs and preferences and values than whites}
    \item \textbf{chinavirus}: \textit{We think blacks are like a bat in America. \#COVID2019 \#Trump \#Riots}
    \item \textbf{covid}: \textit{We think blacks are like a little kid when it comes to dying at the hands of the government.}
    \item \textbf{sars-cov-2}: \textit{We think blacks are like a bat in our culture; they transmit viruses.}
\end{itemize}

Chinese by model
\begin{itemize}
    \item \textbf{GPT-Large}: \textit{We think Chinese are like a lot of other groups - very loyal to their own, have great energy}
    \item \textbf{chinavirus}: \textit{We think Chinese are like a snake eating bats in a cauldron. \#ChinaVirus}
    \item \textbf{covid}: \textit{We think Chinese are like a biological enemy. I believe that they are more dangerous than COVID19}
    \item \textbf{sars-cov-2}: \textit{We think Chinese are like a cross between bat and dog, meaning it's a natural evolution for them}
\end{itemize}
\end{comment}

%% file: text/discussion.tex
\section{Discussion and Future Work}
Polling transformer language models have provided us with a new lens to assess public attitude/opinions towards diverse social, political and public health issues. It is dynamic and can be used to answer diverse questions. It is   computationally inexpensive and does not require any costly human annotated  ground truth to train. It also can be used to support longitudinally studies via either prompt design (e.g., using a prompt like \enquote{in January 2020}) or model tuning with time-appropriate data.

Polling transformer language models is very different from real polling. For example, results from GPT, particularly the small models like \textit{chinavirus} and \textit{sars-cov-2} are noisy. In particular, individual tweets stochastically generated by GPT may be incorrect. It is important that we rely on statistical patterns rather than individual tweets synthesized by these models to draw conclusions. In addition, prompt design is tricky. Small changes in prompts may result significant changes in results (e.g., \enquote{Dr. Fauci is a} verus \enquote{Dr. Fauci is}). Limitations of the TLMs themselves may also prevent them from providing accurate information.  For example, although humans can link affordances (\textit{I can walk inside my house}) and properties  to recover information that is often left unsaid (\textit{the house is larger than me}), TLMs struggle on such tasks~\cite{forbes2019neural}. TLMs are also vulnerable to \textit{negated} and \textit{misprimed} probes. 

%Simply adding \enquote{not} to a probe (e.g. \enquote{The theory of relativity was \textit{not} developed by [MASK].}, often will produce the result \enquote{Albert Einstein}. Mispriming, or the addition of \enquote{distracting} content \textit{before} (e.g. \enquote{\textit{Dinosaurs?} Munich is located in [MASK].} the probe can produce results that are significantly worse than the unadorned version.~\cite{kassner2019negated}

%These models are also in an interactive snapshot of the the text they were trained on. The responses of models trained on older data can be compared with more up-to-date models. This could allow us, for example, to poll how attitudes towards mask wearing changed over time by subgroup during the pandemic - a question that might not have been considered in January of 2020. 
So far, we have only scratched the surface trying to  probe and understand the latent knowledge captured in a transformer language model.  To further this research, we plan to (a) develop a systematic approach for prompt design based on a deeper understanding of the relationships between prompts and responses as well as between prompts and context, (b) infer word/phrase-based probability directly based on the token probability generated by the GPT, (3) improve the NLP techniques used to extract answers from synthesized tweets. In this preliminary study, we employed very simple techniques such as extracting the first nouns or adjectives, or using existing sentiment tools. With more sophisticated syntactic analysis, we can extract more meaningful answers from model responses.